\documentclass[final]{cvpr}

\usepackage{times}
\usepackage{epsfig}
\usepackage{graphicx}
\usepackage{amsmath}
\usepackage{amssymb}

\usepackage[utf8]{inputenc} 
\usepackage[T1]{fontenc}    

\usepackage{hyperref}       
\usepackage{url}            
\usepackage{booktabs}       
\usepackage{amsfonts}       
\usepackage{nicefrac}       
\usepackage{microtype}      

\usepackage{bm}
\usepackage{array}

\usepackage{epsfig}
\usepackage{graphicx}

\usepackage{wrapfig, lipsum}
\usepackage{multirow}
\usepackage{epstopdf}
\usepackage{amssymb}
\usepackage{bbding}

\makeatletter
\newcommand{\namelong}[1]{CondenseNetV2}
\newcommand{\nameshort}[1]{CondenseNetV2}

\newcommand{\namelongm}[1]{CondenseNetV2}
\newcommand{\nameshortm}[1]{CondenseNetV2}

\newcommand{\modulelong}[1]{Learned Sparse Update Module}
\newcommand{\moduleshort}[1]{LSUp Module}

\newcommand{\mG}{G}

\newcommand{\printfnsymbol}[1]{%
  \textsuperscript{\@fnsymbol{#1}}%
}

\def\cvprPaperID{7919} 
\def\confYear{CVPR 2021}

\begin{document}

\title{CondenseNet V2: Sparse Feature Reactivation for Deep Networks}

\author{%
  Le Yang$^{1}$\thanks{Equal contribution.}\ \ \
  Haojun Jiang$^{1}$\footnotemark[1]\ \ \
  Ruojin Cai$^{2}$\ \ \
  Yulin Wang$^{1}$\ \ \
  Shiji Song$^{1}$\ \ \
  Gao Huang$^{1}$\thanks{Corresponding author.}\ \ \
  Qi Tian$^{3}$\\
    $^{1}$Department of Automation, Tsinghua University, Beijing, China\\
    Beijing National Research Center for Information Science and Technology (BNRist),\\
    $^{2}$Cornell University
    $^{3}$Huawei Cloud $\mathbf{\&}$ AI\\
  \texttt{\small \{yangle15, jhj20, wang-yl19\}@mails.tsinghua.edu.cn},    \texttt{\small rc844@cornell.edu},\\
  \texttt{\small\{shijis, gaohuang\}@tsinghua.edu.cn},
   \texttt{\small tian.qi1@huawei.com}
  }

\maketitle
\pagestyle{empty}
\thispagestyle{empty}


\begin{abstract}
    Reusing features in deep networks through dense connectivity is an effective way to achieve high computational efficiency. The recent proposed CondenseNet~\cite{huang2018condense} has shown that this mechanism can be further improved if redundant features are removed. In this paper, we propose an alternative approach named sparse feature reactivation (SFR), aiming at actively increasing the utility of features for reusing. In the proposed network, named \nameshort{}, each layer can simultaneously learn to 1) selectively reuse a set of most important features from preceding layers; and 2) actively update a set of preceding features to increase their utility for later layers. Our experiments show that the proposed models achieve promising performance on image classification (ImageNet and CIFAR) and object detection (MS COCO) in terms of both theoretical efficiency and practical speed.

    \end{abstract} 
\section{Introduction}
Deep convolutional neural networks (CNNs) have achieved remarkable success in the past few years~\cite{vgg, resnet, huang2017densely}. However, their state-of-the-art performance is usually fueled with sufficient computational resources, which hinders deploying deep models on low-compute platforms, e.g., mobile phones and Internet of Things (IoT) products. This issue has motivated a number of researchers on designing efficient CNN architectures~\cite{huang2017densely, chollet2016xception, howard2017mobilenets, zhang2017shufflenet,ma2018shufflenet,sandler2018inverted}. Among these efforts, DenseNet~\cite{huang2017densely} is a promising architecture that improves the computational efficiency by reusing early features with dense connections.
\begin{figure}
	\centering
	\includegraphics[width=0.45 \textwidth]{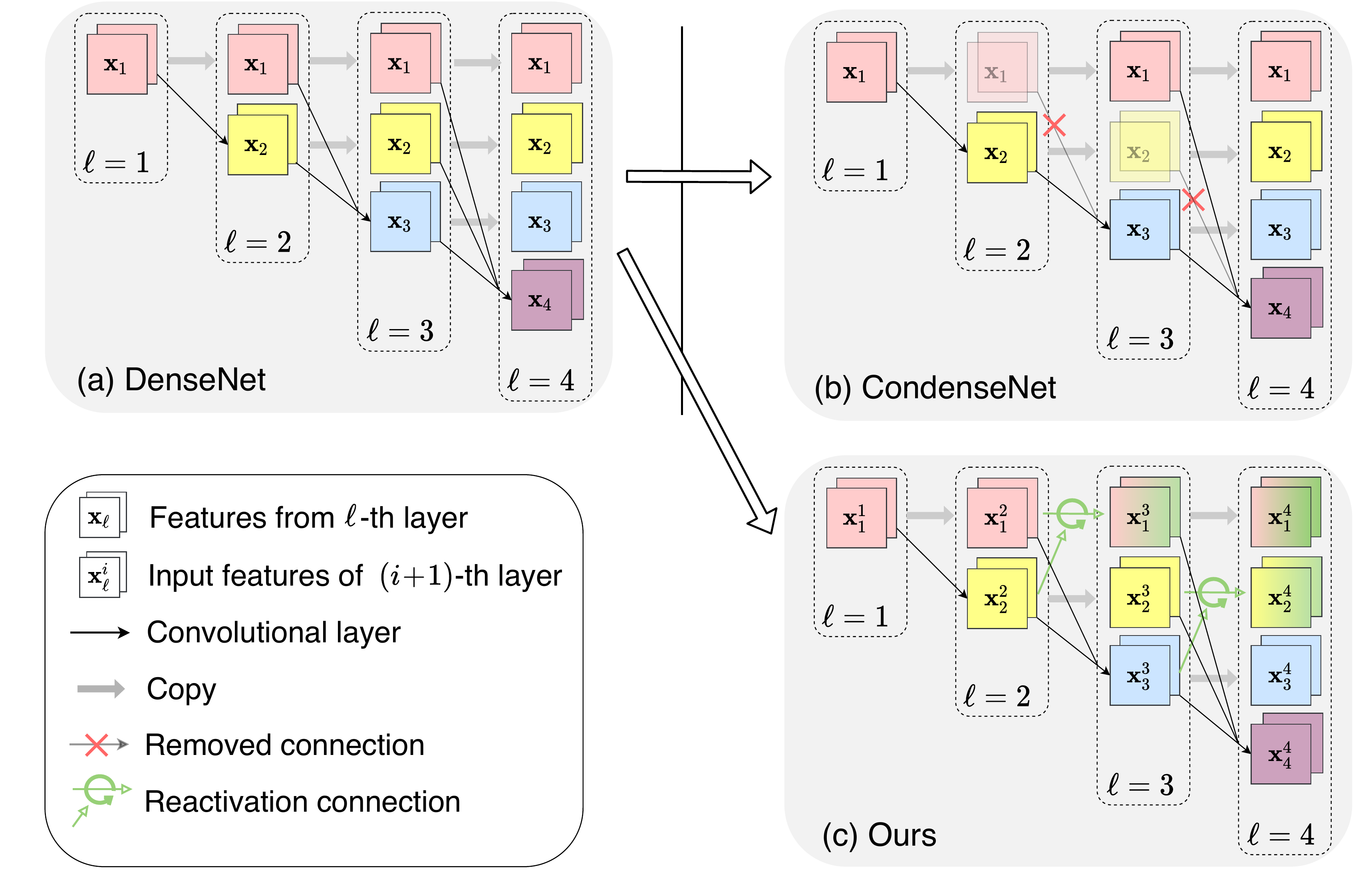}
    \caption{Different feature reuse patterns in (a) DenseNet~\cite{huang2017densely}, (b) CondenseNet~\cite{huang2018condense} and (c) Ours.}
\label{fig1}
\end{figure}

Recently, it has been shown that dense connectivity may introduce a large number of redundancies when the network becomes deeper~\cite{huang2018condense}. In a dense network, the output of a layer will never be modified once it is produced. Given that shallow features will be repeatedly processed by their following layers, directly exploiting them in deep layers might be inefficient or even redundant. CondenseNet~\cite{huang2018condense} alleviates this problem via strategically pruning less important connections in DenseNet. ShuffleNetV2~\cite{ma2018shufflenet} shares a similar spirit, where early features are dropped according to layer-distance, leading to an exponentially decaying of long-distance feature re-usage. Although both models show their effectiveness, we hypothesize that straightforwardly abandoning long connections is overly aggressive. These early features which are considered to be ``obsolete ''  at deeper layers may contain useful information, which can benefit network generalization ability, and potentially contribute to a more efficient model if properly utilized.


In this paper, instead of directly discarding obsolete features, we are interested in whether we can revive them to make obsolete features useful again. To this end, we develop a novel module to conduct \textit{feature reactivation}, which learns to update shallow features and enables them to be more efficiently reused by deep layers. Our main idea is illustrated in Figure \ref{fig1}. Compared to DenseNet~\cite{huang2017densely} and CondenseNet~\cite{huang2018condense}, where earlier features keep unchanged throughout the whole feed-forward process, we propose to allow the outputs of a layer to be reactivated by later layers. Such a way keeps features maps always ``fresh'' at each dense layer, and therefore the redundancy in dense connections can be largely reduced.

Although the \textit{feature reactivation} procedure effectively reduces the redundancy in dense connections, naively reactivating all features will introduce excessive extra computation, which still hurts the overall efficiency. In fact, it is unnecessary to reactivate all features since a large number of them can be already effectively reused without any change in dense connections, resulting in that only \textit{sparse feature reactivation} (SFR) is required. For this purpose, we develop a cost-efficient SFR module which actively and selectively \textit{reactivates} early features at each layer, using the increments learned from the newly produced feature maps. Importantly, both the features to be updated and the updating formulas are determined \textit{automatically} via learning. During the training process, we first assume all previous features require reactivating, and then gradually remove the reactivation that have less effect on feature re-usage. Moreover, the resulting SFR modules can be converted to efficient group convolutions at test time. As a consequence, the proposed method involves minimal extra computational cost or latency and keeps early features ``fresh'' even through very deep layers, which leads to a significant efficiency gain.



We implement SFR on the basis of the efficient CondenseNet~\cite{huang2018condense}, where the SFR along with the learned group convolutions (LGCs)~\cite{huang2018condense} can be learned compatibly to improve the efficiency of dense networks. The resulting network, CondenseNetV2, are empirically evaluated on image classification benchmarks (ImageNet and CIFAR) and the COCO object detection task. The results demonstrate that SFR significantly boosts the performance by encouraging long-distance feature reusing, and that CondenseNetV2 compares favorably with even state-of-the-art light-weighted deep models. We also show that SFR can be plugged into any CNNs that adopt the concatenation based feature reusing mechanism to further improve their efficiency, such as ShuffleNetV2~\cite{ma2018shufflenet}.

\section{Related Work}

\paragraph{Efficient network architectures.}
Designing better network architectures is an effective way to improve the computation efficiency of deep networks. Efficient building units are introduced for light-weighted CNN architectures. For instance, MobileNets~\cite{howard2017mobilenets,sandler2018inverted,howard2019searching} propose inverted residuals and linear bottlenecks to build network architectures. Sandglass blocks, which flip the inverted residuals, are developed in MobileNeXts~\cite{zhou2020rethinking}.  Cheap
operations are developed for generating features in GhostNet~\cite{ghostnet}. In addition, shuffle layer and learned group convolution (LGC) are employed by ShuffleNets~\cite{zhang2017shufflenet,ma2018shufflenet} and CondenseNet~\cite{huang2018condense}, respectively. Recent study also shows that developing dynamic neural networks~\cite{han2021dynamic} can obviously improve the efficiency of deep models, such as \cite{huang2018multi,yang2020resolution,wang2020nips}. In this paper, we follow the first line of the research and propose a novel efficient unit named SFR module. The proposed deep models with SFR module retains the simplicity of CondenseNet while significantly improves its accuracy on image classification and detection tasks for mobile applications.

\vspace{-13pt}
\paragraph{Densely connected neural network.}
Compared to ResNet~\cite{resnet} and it variants~\cite{resnext,zagoruyko2016wide}, DenseNet architectures~\cite{huang2017densely,huang2018condense,huang2018multi,yang2020resolution} can achieve a higher computational efficiency by encouraging feature reuse. However, superfluous re-usage may introduce redundant connections. To address this problem, existing work mainly proposes to remove dense connections to feature maps that are less useful~\cite{huang2018condense,wang2019FLGC}, or to discard long-range connections according to a predefined probability~\cite{ma2018shufflenet}. However, as these seemingly redundant connections may have large potential in deep layers if properly utilized, we propose to conduct sparse feature reactivation to deal with the redundant connections rather than pruning them.


\vspace{-6pt}
\paragraph{Filter pruning.}
Although the sparsifying procedure in SFR module is related to filter pruning methods~\cite{Wang2019Pruning,he2017channel,li2016pruning,he2018pruning,liu2017learning}, our method differs greatly from filter pruning methods in the way dealing with the redundant connections. Instead of removing connections between layers where the feature re-usage is superfluous, our approach aims at building reactivation connections to revive obsolete features to increase their utility. Notably, our reactivation idea is orthogonal to filter pruning, and both are utilized for building \nameshort{}. Additionally, compared with recent work~\cite{Meng_2020_CVPR}, which proposes to graft new weights to the unimportant filters, the proposed SFR module reactivates obsolete features to improve efficiency.

\section{Method}

A recently confirmed inefficiency in DenseNet~\cite{huang2017densely} architecture lies in the presence of long-distance connections~\cite{huang2018condense}, where the deeper layers seem to consider the early features as ``{obsolete}'' ones and ignore them during learning new representations. CondenseNet~\cite{huang2018condense} and ShuffleNetV2~\cite{ma2018shufflenet} alleviate this inefficiency through strategically pruning redundant connections and exponentially discarding cross-layer connections, respectively. In this paper, we postulate in this paper that directly abandoning shallow features can be an overly aggressive design. To be specific, we find that by involving a learnable \textit{sparse feature reactivation} (SFR) module with a negligible computational cost at each layer, the originally ``{obsolete}'' features can be  ``{reactivated}'' and hence effectively exploited by the later layers. In this section, we first describe the details of the proposed SFR module, and then implement it to build our light-weighted networks.



\subsection{Sparse Feature Reactivation}
\paragraph{Feature reuse mechanism.}
We first formulate the feature reusing mechanism introduced in \cite{huang2017densely}. Assume that a standard network block of $L$ layers produces $L$ feature maps $\mathbf{x}_1, \mathbf{x}_2, \ldots, \mathbf{x}_{L}$, where $\mathbf{x}_{\ell}$ is the output of the $\ell$-th layer and $\mathbf{x}_0$ denotes the input feature. Since all previous layers are connected to the $\ell$-th layer via the dense connections, the composite function $H_{\ell}(\cdot)$ of the $\ell$-th layer will take all of $\mathbf{x}_0, \ldots, \mathbf{x}_{\ell-1}$ as inputs:
\begin{equation}
\label{dense}
    \mathbf{x}_{\ell} = H_{\ell}([\mathbf{x}_0, \mathbf{x}_1, \ldots, \mathbf{x}_{\ell-1}]).
\end{equation}
In CondenseNet~\cite{huang2018condense}, learned group convolutions (LGC) are employed in $H_{\ell}(\cdot)$ to automatically learn the input groupings and remove unimportant connections, while in ShuffleNetV2~\cite{ma2018shufflenet}, inputs of $H_{\ell}$ will be dropped according to their distance to layer $\ell$. Consequently, both of them remove the superfluous long-distance connections between layers, which are proven effective in term of efficiency. However, given that the output $\mathbf{x}_{\ell}$ will never change once it is produced, a side effect is that these seemingly less useful features from shallow layers tend to be permanently discarded by deeper layers. This \emph{static} design may impede exploring more efficient feature reusing mechanisms. To this end, we propose a cost-efficient \emph{sparse feature reactivation} module, enabling obsolete features to be cheaply revived.

\vspace{-5pt}
\paragraph{Reactivating obsolete features.}
We start by describing the details of feature reactivation. For $\ell$-th layer, we introduce a reactivation module denoted by $\mG_{\ell}(\cdot)$. The module takes $\mathbf{x}_{\ell}$ as input, and its output $\mathbf{y}_{\ell}$ is used to reactive features from preceding layers. In this paper, we define the reactivation operation, $U(\cdot,\cdot)$, as adding\footnote{Other reactivation schemes may also be considered, such as applying channel-wise attention or spatial attention. However, in this paper, we note that a straightforward sum has already achieved good performance.} the increment  $\mathbf{y}_{\ell}$. A dense layer with feature reactivation can be written as
\begin{align}
\label{eq-up1}
	& \mathbf{x}^{\text{in}}_{\ell}\leftarrow[\mathbf{x}_0, \mathbf{x}_1, \ldots, \mathbf{x}_{\ell-1}], \quad \mathbf{x}_{\ell}\!=\!H_{\ell}(\mathbf{x}^{\text{in}}_{\ell}), \\
\label{eq-up2}
	& \mathbf{y}_{\ell}\!=\!\mG_{\ell}(\mathbf{x}_{\ell}), \quad \mathbf{x}^{\text{out}}_{\ell}\!=\!U(\mathbf{x}^{\text{in}}_{\ell}, \mathbf{y}_{\ell}), \\
	& [\mathbf{x}_0, \mathbf{x}_1, \ldots, \mathbf{x}_{\ell-1}]\leftarrow\mathbf{x}^{\text{out}}_{\ell},
\end{align}
where $\mathbf{x}^{\text{out}}_{\ell}$ is the reactivated output feature. With $H_{\ell}(\cdot)$, the $\ell$-th layer learns to produce new feature $\mathbf{x}_{\ell}$. Additionally, previous representations ($\mathbf{{x}}_{i}, i\!=\!1,..\ell\!\!-\!\!1$) will be reactivated to increase their utility.

\begin{figure}[htp]
	\vspace{-14pt}
    \centering
	\includegraphics[width=0.5 \textwidth]{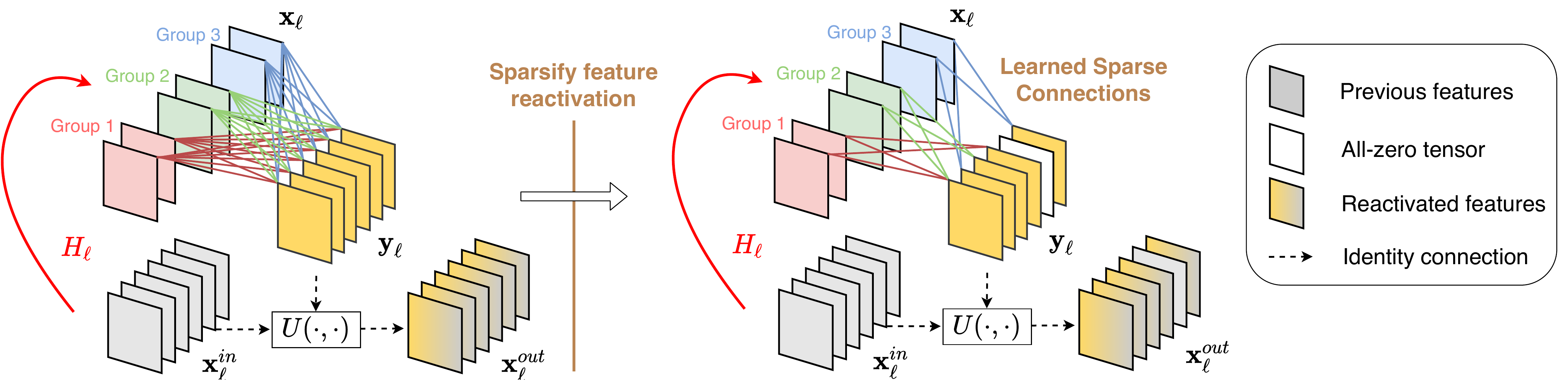}
	\caption{Sparsify the feature reactivation.}
	\label{fig:tlgc-1}
\end{figure}

Obviously, it is unnecessary to reactivate all features since a large number of them can be effectively reused without any change (shown in DenseNet~\cite{huang2017densely}). We also empirically observe that dense reactivation will introduce much computation and degrade the overall efficiency of the network. Therefore, we seek to automatically find the features required to be reactivated and merely refresh them. In the following, this aim is formulated by a pruning based approach and can be achieved gradually during training (shown in Figure \ref{fig:tlgc-1}). The resulting architecture is named as sparse feature reactivation (SFR) module.


\paragraph{Sparse feature reactivation (SFR).}
WLOG, we assume that the reactivation module $\mG_{\ell}(\cdot)$ consists of a regular 1$\times$1 convolutional layer followed by batch normalization (BN)~\cite{batch-norm} and a rectified linear unit (ReLU)~\cite{relu}. The size of the filter weight matrix $\mathbf{F}$ in $\mG_{\ell}(\cdot)$ is represented as $(O, I)$, where $O$ and $I$ denote the number of output and input channels\footnote{We can apply max-pooling on the absolute value of the 4D weights, $\mathbf{F} \in \mathbb{R}^{O\times I\times k \times k}$,  to generate the matrix with the size of $(O, I)$ when dealing with larger convolutional kernels.}. In each $\mG_{\ell}(\cdot)$, we divide $\mathbf{x}_\ell$ into $G$ groups, and the $\mathbf{F}$ is split correspondingly among the input channel dimension to obtain $G$ groups $\mathbf{F}^1, \ldots, \mathbf{F}^G$, and each of them has the size of $(O, I/G)$. To sparsify the reactivation connections, we further define a sparse factor $S$, which may differ from $G$, and allow each group to only select $\frac{O}{S}$ output channels to reactivate after training.

During training, in each $\mG_{\ell}(\cdot)$, the connection pattern is controlled by $G$ binary masks. Therefore, we aim at learning these binary masks $\mathbf{M}^g \in\{0,1\}^{O\times \frac{I}{G}}, g \!=\!1,...,G$  to screen out unnecessary connections in $\mathbf{F}^g$ by zeroing the corresponding values. In other words, the weight of $g$-th group can be obtained by $\mathbf{M}^g \odot \mathbf{F}^g$, where $\odot$ denotes the element-wise multiplication.

We then introduce how to train a network with SFR modules in an end-to-end manner.
Inspired by~\cite{huang2018condense}, the whole training process consists of $S - 1$ sparsification stages followed by an optimization stage. Assuming that $E$ denotes the total number of training epochs, we set the training epochs of each sparsification stage to $\frac{E}{2(S-1)}$ and optimization stage to $\frac{E}{2}$. During training, the SFR module first reactivates all features, and then gradually removes the superfluous connections. Therefore, at the beginning of training, we set all $\mathbf{M}^g $ to all-ones matrices, thus all input feature maps in $g$-th group are connected with all output features. During sparsification, the importance of reactivating $i$-th output within $g$-th group is measured by the L1-norm of the corresponding weights $\sum_{j=1}^{I/G} |\mathbf{F}_{i,j}^g|$. At the end of each sparsification stage, $\frac{O}{S}$ output features whose L1-norms are smaller than others are pruned in $g$-th group, and $\mathbf{M}_{i, j}^g$ is set to zero for all $j$ in $g$-th group for each pruned output feature map $i$. Note that, if $i$-th output feature map is pruned from every input group, the $i$-th feature map in $\mathbf{y}$ will equal to $\mathbf{0}$, which implies that the $i$-th feature map from previous layers do not need to reactivate. Therefore, after training, each input group will only update the outputs with a portion of $1/S$, which means that for final $\mathbf{M}^g$, we have \begin{small}$\sum_{i=1}^O \mathbf{M}^g_{i,j}\!= \!\frac{O}{S} $\end{small}. The higher the value of $S$, the sparser the connection pattern is.

\begin{figure}[htp]
	\centering
	\includegraphics[width=0.47 \textwidth]{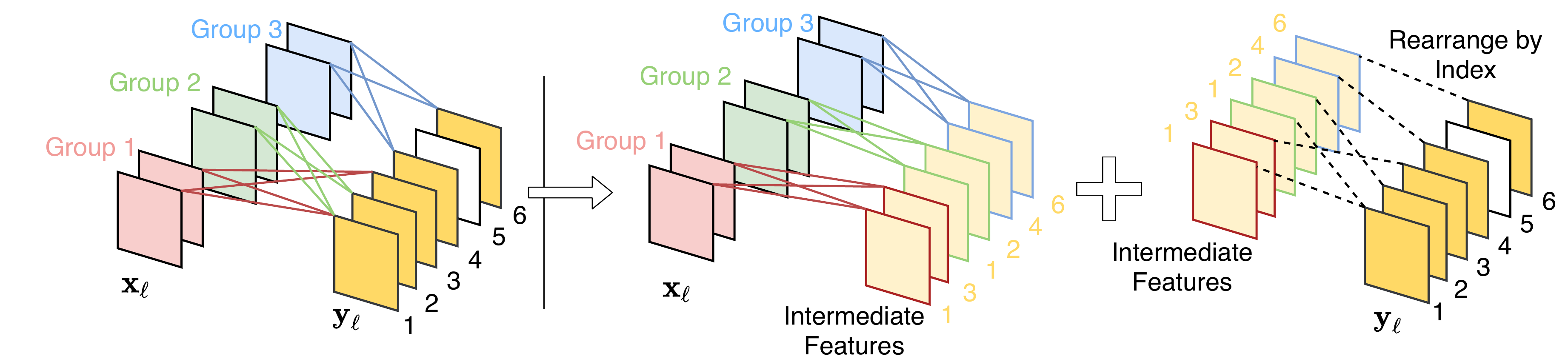}
	\caption{Convert to standard group convolution ($S\!=\!3$ and $G\!=\!3$). }
	\label{fig:tlgc-2}
\end{figure}

\paragraph{Convert to standard group convolution.} At test time, our SFR model can be implemented
using a standard group convolution and an index layer, allowing for efficient computation in practice. This is illustrated in Figure~\ref{fig:tlgc-2}: the converted group convolution contains $G$ groups with output and input channels as $(\frac{OG}{S}, I)$. After generating the intermediate features with the group convolution, the index layer is applied to rearrange the features by their indices to obtain the $\mathbf{y}_{\ell}$. Note that the intermediate features with the same index will first be summed and then arranged according to the index.

\vspace{-10pt}
\subsection{Architecture Design}
\begin{figure}[htp]
    \vspace{-11pt}
	\centering
	\includegraphics[width=0.45\textwidth]{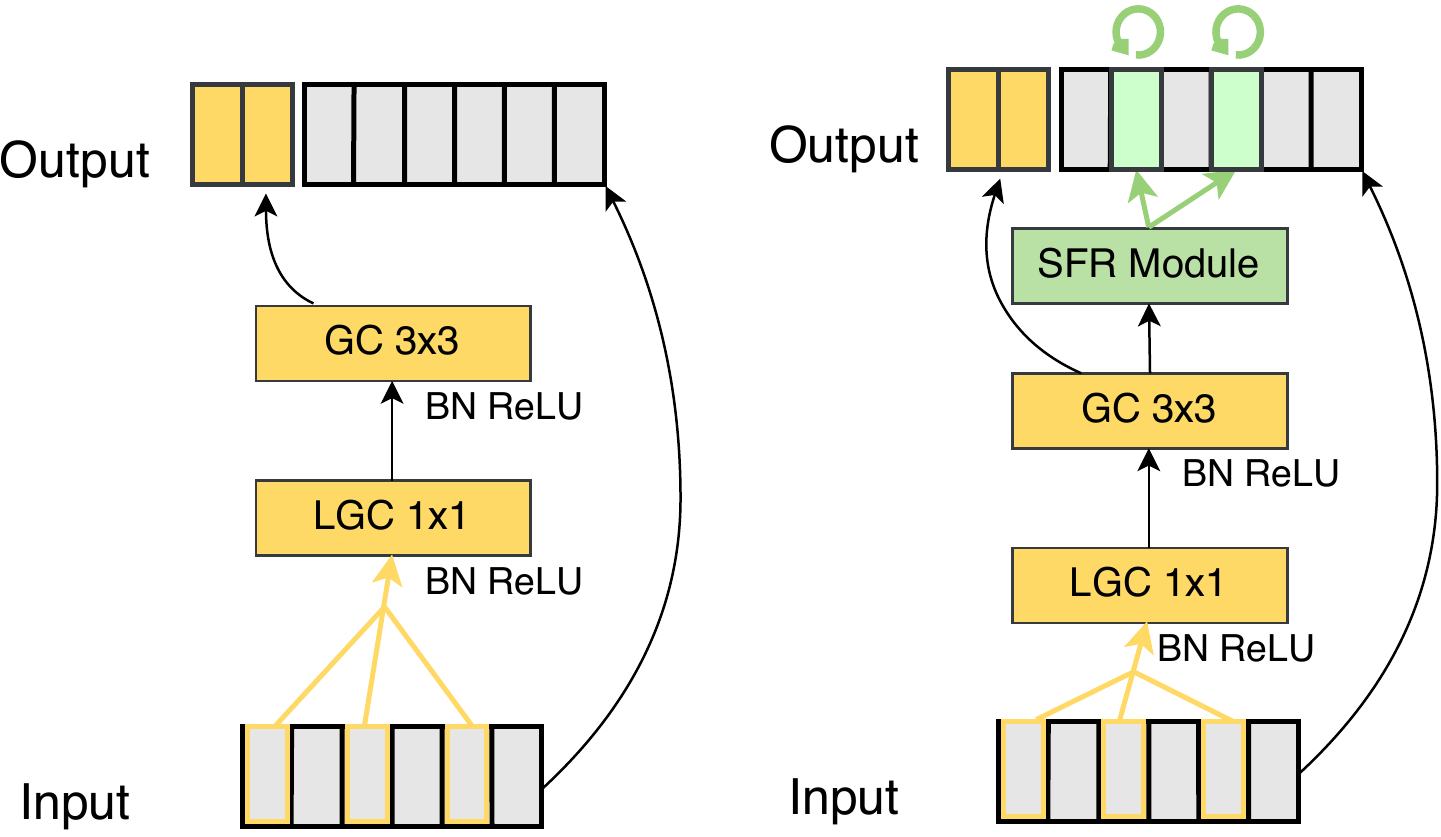}
	\caption{A dense layer in CondenseNet (left), and \nameshort{} (right). (LGC: learned group convolution; GC: group convolution)}
	\label{fig:condensev2block}
\end{figure}

\paragraph{Architecture of CondenseNetV2.}
Based on the proposed SFR module, the new dense layer of CondenseNetV2 is shown in Figure \ref{fig:condensev2block} (right), which is designed on the basis of CondenseNet~\cite{huang2018condense}. In the proposed architecture, the LGC first selects important connections and the new representations $\mathbf{x}_{\ell}$ are generated based on these selected features using Eq. (\ref{eq-up1}). Then, the SFR module takes $\mathbf{x}_{\ell}$ as input and learns to reactivate the obsolete representations. The refreshed features can be derived by Eq. (\ref{eq-up2}). Following \cite{huang2018condense}, we shuffle the output channels of each convolutional layer to ensure communication between different groups. It is worth knowing that the \nameshort{} is essentially different from CondenseNet~\cite{huang2018condense}: the outputs of each layer in CondenseNet~\cite{huang2018condense} will never change once it is produced. Therefore, the potential re-usage of previous features can be blocked. In contrast, old features can be reactivated in each layer of \nameshortm{}, resulting in a more efficient and effective feature reuse mechanism.

The architecture of \nameshort{} follows the exponentially increasing growth rate and fully dense connectivity design principle~\cite{huang2018condense}. Based on the newly designed SFR-DenseLayer, we develop our \namelongm{} as presented in Table \ref{tab:ArchiImageNet}. The squeeze and excite (SE) module~\cite{hu2017squeeze} and hard-swish nonlinearity function (HS) are also applied following \cite{howard2019searching}. The presented architecture provides a basic design for reference, further hyper-parameters tuning or network architecture searching can further boost the performance.

\begin{table}\footnotesize
\caption{Network architecture of \nameshortm{}. The number of layers and the growth rate of $i$-th dense block are denoted by $d_i$ and $k_i$, respectively. SE and HS denote whether using SE and HS module in this block.}
\renewcommand{\arraystretch}{1.1}
\begin{center}
\begin{tabular}{c||c|c|c|c|c}
\hline
Input & Operator & $d$ & $k$ &SE &HS \\
\hline
$224\!\times\!224$ &  Conv2d $3\!\times\!3$ (stride 2)& -& -& - & -\\
\hline
$112\!\times\!112$ & SFR-DenseLayer & $d_1$ & $k_1$& - & -   \\
$112\!\times\!112$ & AvgPool $2\!\times\!2$ (stide 2)&- & - & - & -\\
\hline
$56\!\times\!56$ & SFR-DenseLayer &$d_2$ & $k_2$ &-& -\\
$56\!\times\!56$ & AvgPool $2\!\times\!2$ (stride 2)& -&- &-  &-  \\
\hline
$28\!\times\!28$ & SFR-DenseLayer &$d_3$ & $k_3$ & -& 1\\
$28\!\times\!28$ & AvgPool $2\!\times\!2$ (stride 2)  & -&- &- &- \\
\hline
$14\!\times\!14$ & SFR-DenseLayer & $d_4$ & $k_4$ & 1 & 1 \\
$14\!\times\!14$ & AvgPool $2\!\times\!2$ (stride 2) & -& -& -&  -\\
\hline
$7\!\times\!7$ & SFR-DenseLayer &$d_5$ & $k_5$ &1& 1\\
\hline
$1\!\times\!1$ &  AvgPool $7\!\times\!7$ &-& -&- & - \\
\hline
$1\!\times\!1$ & Conv2d $1\!\times\!1$ &- &- & 1& -\\
\hline
$1\!\times\!1$ & FC &  - & -& -& -\\
\hline
\end{tabular}
\end{center}
\vspace{-5pt}
\label{tab:ArchiImageNet}
\end{table}

\subsection{Sparse Feature Reactivation in ShuffleNetV2}
\begin{figure}[htp]
	\centering
	\includegraphics[width=0.45 \textwidth]{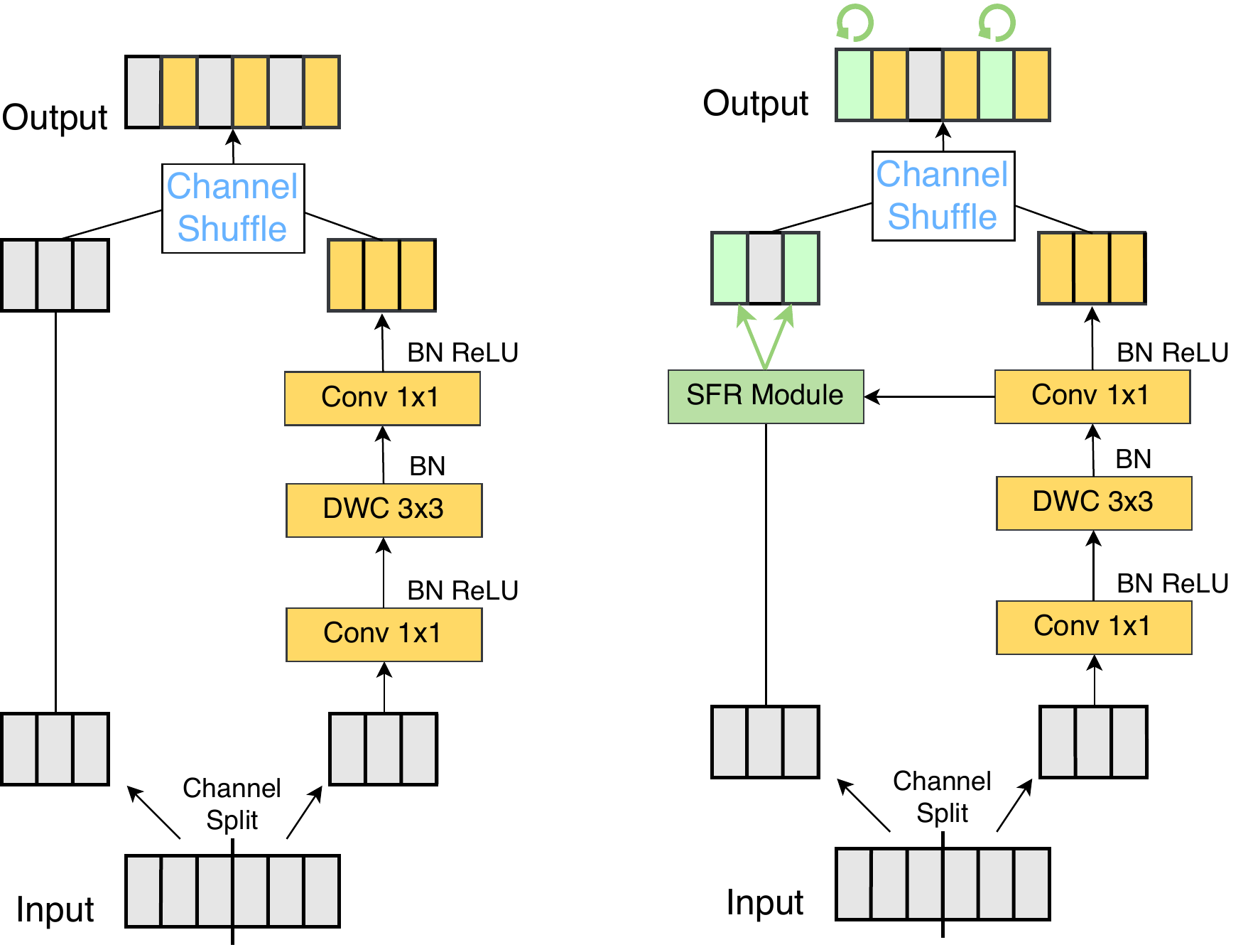}
	\caption{A building unit in ShuffleNetV2 (left), and the ShuffleNetV2 implemented with SFR module (right). (DWC: depth-wise convlution)}
	\label{fig:shufllev2block}
    \vspace{-5pt}
\end{figure}

As SFR is able to be plugged into any CNNs with the feature reusing mechanism, we claim that ShuffleNetV2~\cite{ma2018shufflenet} can also benefit from the proposed SFR module. The implementation details are illustrated in Figure \ref{fig:shufllev2block}. We refer to the modified ShuffleNetV2 with sparse feature reactivation as SFR-ShuffleNetV2. Note that in SFR-ShuffleNetV2, only basic units conduct the feature reactivation, and the units for spatial down sampling keep unchanged. The detailed architecture is provided in the Appendix.

\section{Experiments}
\vspace{-5pt}
We empirically demonstrate the effectiveness of the proposed SFR module and \nameshort{} on image classification and object detection tasks, and compare with state-of-the-art light-weighted CNN architectures. Code is available at \url{https://github.com/jianghaojun/CondenseNetV2}.
\vspace{-8pt}
\paragraph{Dataset.}
Experiments are conducted on several benchmark visual datasets, including CIFAR-10 and CIFAR-100~\cite{cifar}, ImageNet (ILSVRC2012~\cite{deng2009imagenet}), and MS COCO object detection benchmark~\cite{lin2014microsoft}.

The CIFAR-10 and CIFAR-100 datasets consist of $32\!\times\!32$ RGB images, with 10 and 100 classes of natural scene objects, respectively. Both CIFAR datasets contain 50,000 training images and 10,000 test images. On the two CIFAR datasets, following \cite{huang2017densely}, we apply a set of transformations to augment the training set. The ImageNet (ILSVRC2012~\cite{deng2009imagenet}) classification dataset contains 1.2 million training images and 50,000 validation images, with 1000 classes. Data augmentation schemes are applied at training and we adopt a $224\!\times\!224$ center crop at test time. On MS COCO dataset, following \cite{ren2015fasterRCNN,lin2017featurePyramid}, we use the \textit{trainval35k} split as training data and report the results in mean Average Precision (mAP) on \textit{minival split} with 5000 images.


\subsection{Efficiency of Sparse Feature Reactivation}
\vspace{-5pt}
In this subsection, we conduct a series of experiments on densely connected networks with sparse feature reactivation to verify its effectiveness.
\vspace{-5pt}
\paragraph{Reactivated features.}

\begin{figure}
\centering
\includegraphics[width=0.48\textwidth]{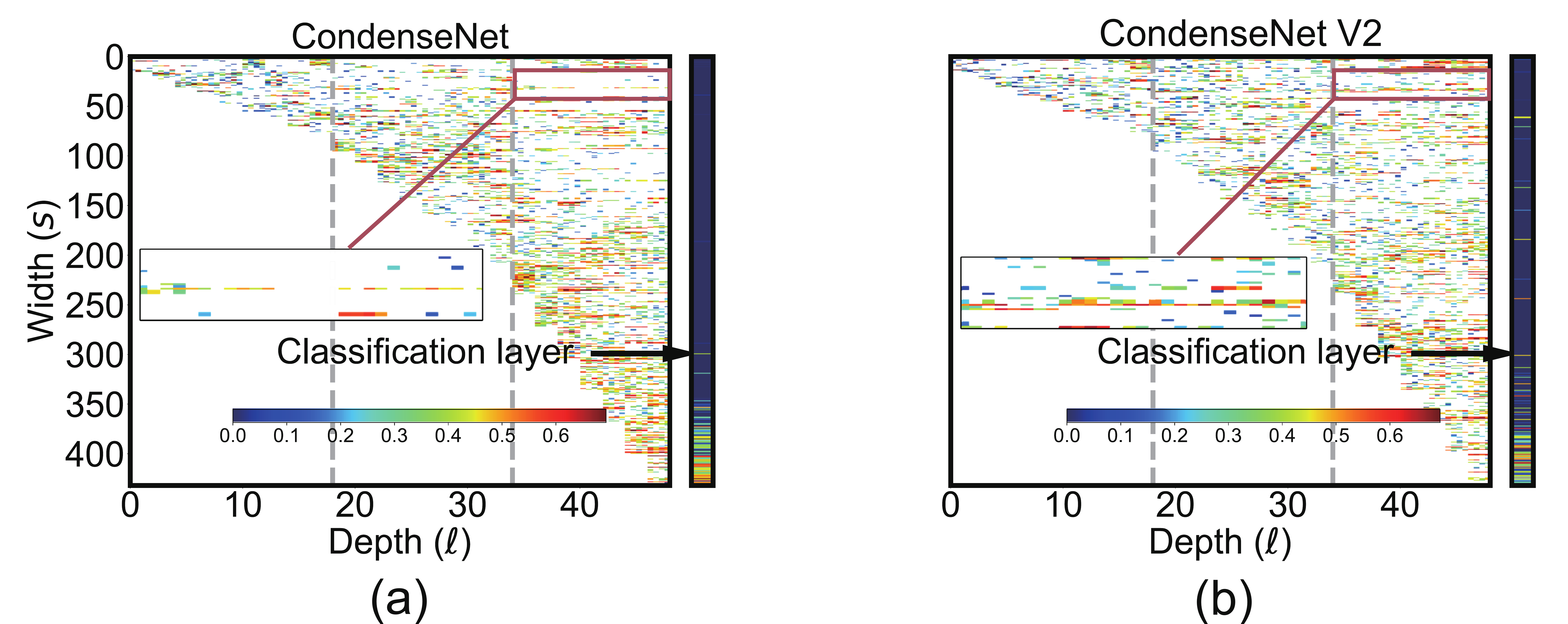}
\caption{Norm of weights between layers per filter group of CondenseNet and \nameshort{} trained on CIFAR-10.}
\label{fig:weight-LGC}
\vspace{-10pt}	
\end{figure}

\begin{figure*}
\centering
\includegraphics[width=1\textwidth]{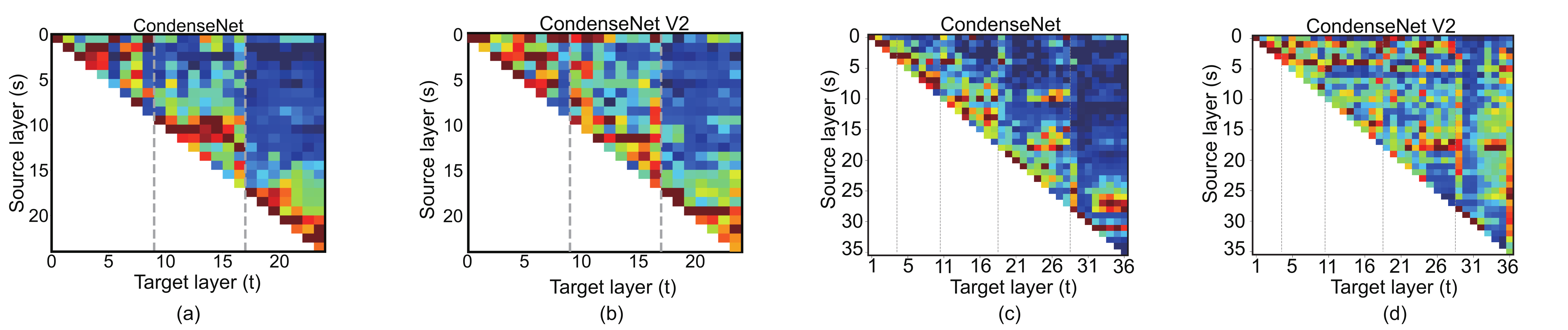}
\caption{Norm of weights between layers per filter block of CondenseNet ((a) and (c)) and \nameshort{} ((b) and (d)). The models are trained on CIFAR-10 ((a) and (b)) and ImageNet ((c) and (d)).}
\label{fig:weight-LGC-layer}
\vspace{-5pt}	
\end{figure*}

As the proposed SFR module is designed for reactivating the redundant features to improve the network efficiency, a natural question is whether the utility of these obsolete features indeed shows great importance at later layers after being reactivated. To investigate this question, Figure~\ref{fig:weight-LGC} and \ref{fig:weight-LGC-layer} visualize the learned weights for CondenseNet~\cite{huang2018condense} and \nameshort{} to verify whether those reactivation will indeed encourage feature reuse.

Figure~\ref{fig:weight-LGC} shows detailed weight strength (averaged absolute value of non-pruned weights) between a filter group of a certain layer (corresponding to a column in the figure) and an input feature map (corresponding to a row in the figure). For each layer, there are four filter groups (consecutive columns). Red dots are connections with significant contributions and white dots are connections that have been pruned. One can observe that connections in CondenseNet are more concentrated in neighbor layers, while long-distance connections appear more frequently in \nameshort{} (shown by dense colored dots in the top-right corner). This implies that later layers make more use of feature maps produced by early layers in \nameshort{} than in CondenseNet.

Figure~\ref{fig:weight-LGC-layer} shows the overall connection strength between two layers in the CondenseNet and \nameshort{}. The color shows the L1-norm of weights between layers and red means large weight. We notice that the top-right parts of figures for \nameshort{} are more brilliant than these corresponding areas of figures for CondenseNet, which implies that the utility of early features largely increases in \nameshort{}. As the performance of \nameshort{} is shown to be superior to CondenseNet, we can conclude that the dense network will benefit from sparse feature reactivation. This validates our hypothesize: although the features produced by early layers seem unimportant at deep layers in dense networks, they may have potential after being reactivated. Moreover, from the results in Figure~\ref{fig:weight-LGC-layer} in (b) and (d), we further observe that early features are more frequently utilized at deep layers in the model trained on ImageNet than the model trained on CIFAR-10, from which we can infer that the reactivated early features show more importance in complicate tasks.

\vspace{-5pt}
\paragraph{Necessity of sparse feature reactivation.}

\begin{figure}
	\centering
	\includegraphics[width=0.5 \textwidth]{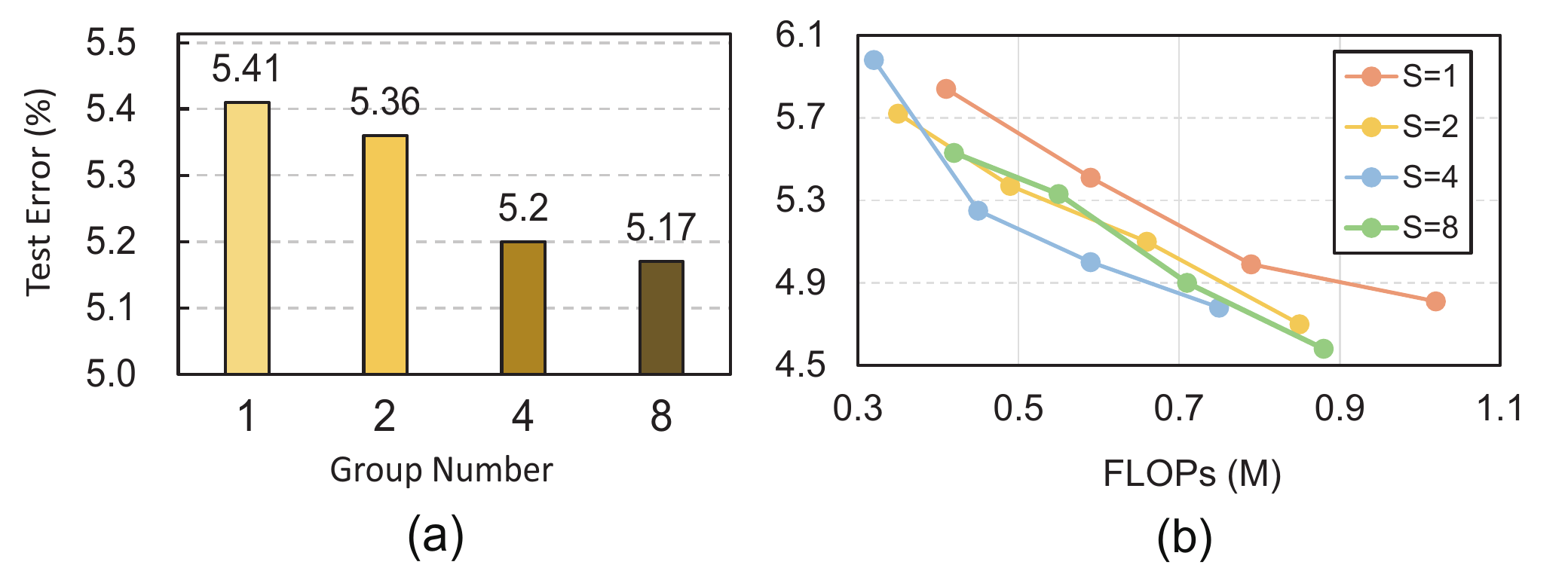}
\caption{(a): SFR module with different group number (G) in \nameshort{}.  (b): \nameshort{} with different sparse factor (S) for SFR module. }
\vspace{-5pt}
	\label{fig:ablation}
\end{figure}

Although the dense feature reactivation can improve the network performance, the involved extra computation is much, which degrades the overall efficiency of the network. Therefore, it is necessary to make the reactivation sparse. We conduct experiments on CIFAR-10 using \nameshortm{} with $S=1$ and $S>1$. The experimental results are shown in Figure \ref{fig:ablation} (b). From the results, we observe that the dense feature reactivation ($S=1$) becomes inefficient due to the heavy extra computational overheads. On the contrary, conducting sparse feature reactivation ($S>1$) can deal with the aforementioned problems effectively, and therefore, can boost the network performance with minor extra computation.

\vspace{-5pt}
\paragraph{The hyper-parameters in SFR module.}	
We conduct the experiments with \nameshort{} on CIFAR-10 for evaluation. We fix the group number and the condense factor for LGC layers, and we only compare different settings for SFR module. In Figure~\ref{fig:ablation} (a), the sparse factor is fixed to 4 and we show the effect of group number $G$ which actually does not affect the FLOPs of the network. One can observe that as $G$ increases, the performance improves gradually. This is due to that a finer-grained sparsification, which corresponds to a large $G$, is usually able to achieve higher efficiency. In Figure~\ref{fig:ablation} (b), we compare \nameshort{}s with varying sparse factors in SFR module. A network with a sparse factor $S$ of 1 means that all reactivation connections are preserved (dense reactivation). If $S$ is increased to $4$, then each layer only keeps a quarter of reactivation connections. The results show that $S=4$ outperforms other settings. Also, all the settings with $S>1$ perform better than the setting with $S=1$, indicates that removing a set of unnecessary connections is indeed important for building efficient \nameshort{}.

\subsection{Experiments on ImageNet}
\label{sec:imagenet}
\paragraph{Implementation details.}
Following the common practice \cite{howard2017mobilenets,zhang2017shufflenet,ma2018shufflenet}, the proposed network has three levels of computational complexity. The \nameshortm{} with different sizes are summarized in Table~\ref{tab:configImageNet}, where $d$ and $k$ are number of layers and growth rate of each dense block, respectively. In all experiments, we use the same condense factor ($C$), group number ($G$), and sparse factor ($S$) for all LGCs and SFR modules in the network. \nameshort{} are trained using the stochastic gradient descent (SGD) optimizer with an initial learning rate of 0.4, the cosine learning rate~\cite{loshchilov2016sgdr}, and The batch size of 1024. To compare with SOTA baselines, we implement an Augmented Setting differing from the original setting in~\cite{huang2018condense}. More details can be found in Appendix C.

\begin{table}\small
    \caption{Network configurations for ImageNet models.}
    \label{tab:configImageNet}
    \centering
    \begin{tabular}{c|c|c|c}
    \toprule
        Setting & $\{d\}$ & $\{k\}$ & $C,S,G$\\
    \midrule
        \nameshortm{}-A & 1-1-4-6-8 & 8-8-16-32-64 & 8\\
        \nameshortm{}-B & 2-4-6-8-6 & 6-12-24-48-96& 6\\
        \nameshortm{}-C & 4-6-8-10-8 & 8-16-32-64-128 & 8 \\
    \bottomrule
    \end{tabular}
\end{table}

\vspace{-5pt}
\paragraph{Results on ImageNet.}
We conduct experiments on ImageNet to evaluate the effectiveness of the proposed methods. As the SFR module can be deployed in both CondenseNet and ShuffleNetV2, we first compare the original model and the network with SFR module in Table \ref{tab:ImageNetResult1}. From the results, we can see that the latter clearly exceeds the former. The computational cost of \nameshort{}-A is 18\% lower than CondenseNet-A (46M v.s. 56M). Using SFR on ShuffleNet can also boost the efficiency with minor extra FLOPs (fewer than 3\%). Moreover, we conduct ablation studies on ImageNet to show how each additional design benefit the original CondenseNet. The results are shown in Table \ref{tab:ablation}, from which we observe that implementing the proposed SFR can boost the performance of CondenseNet by a large margin.

\begin{table}\footnotesize
    \caption{Top-1 and Top-5 classification error rate (\%) on ImageNet.}
    \begin{center}%
    \begin{tabular}{p{2.9cm}|p{0.8cm}<{\centering}p{1.3cm}<{\centering}  p{1.3cm}<{\centering}}
    \toprule
    Model & FLOPs &  Top-1 err. & Top-5 err. \\
    \midrule
    CondenseNet-A~\cite{huang2018condense} & 56M &  43.5 & 20.2 \\
    CondenseNetV2-A & 46M & 35.6 & 15.8\\
    CondenseNet-B~\cite{huang2018condense}& 132M & 33.9 & 13.1\\
    CondenseNetV2-B & 146M & 28.1 & 9.7\\
    \midrule
    ShuffleNetV2 1.0$\times$~\cite{ma2018shufflenet} & 146M & 30.6 & 11.1\\
    SFR-ShuffleNetV2 1.0$\times$ & 150M & 29.9 & 10.9\\
    ShuffleNetV2 1.5$\times$~\cite{ma2018shufflenet} & 299M & 27.4 & 9.4\\
    SFR-ShuffleNetV2 1.5$\times$& 306M & 26.5 & 8.6\\
    \bottomrule
    \end{tabular}%
    \end{center}
    \label{tab:ImageNetResult1}
    \vspace{-15pt}	
\end{table}

\begin{table}[h]\footnotesize
    \caption{Ablations of CondenseNetV2-A on ImageNet.}
    \begin{center}%
    \begin{tabular}{p{2.2cm} | p{0.4cm}<{\centering} p{0.4cm}<{\centering} p{1.5cm}<{\centering} | p{1.3cm}<{\centering}}
    \toprule
    Model & SFR & SE/HS & Augmented Setting & Top-1 err. \\
    \midrule
    CondenseNet~\cite{huang2018condense} &  &  &  & 43.5 \\
    CondenseNetV2 & \checkmark &  &  & 39.8(\textcolor{blue}{$\downarrow$3.7}) \\
    CondenseNetV2 & \checkmark & \checkmark &  & 37.5(\textcolor{blue}{$\downarrow$6.0}) \\
    CondenseNetV2 & \checkmark & \checkmark & \checkmark & 35.6(\textcolor{blue}{$\downarrow$7.9}) \\
    \bottomrule
    \end{tabular}
    \end{center}
    \label{tab:ablation}
    \vspace{-10pt}	
\end{table}

\begin{figure*}\small
\centering
	\includegraphics[width=1.0 \textwidth]{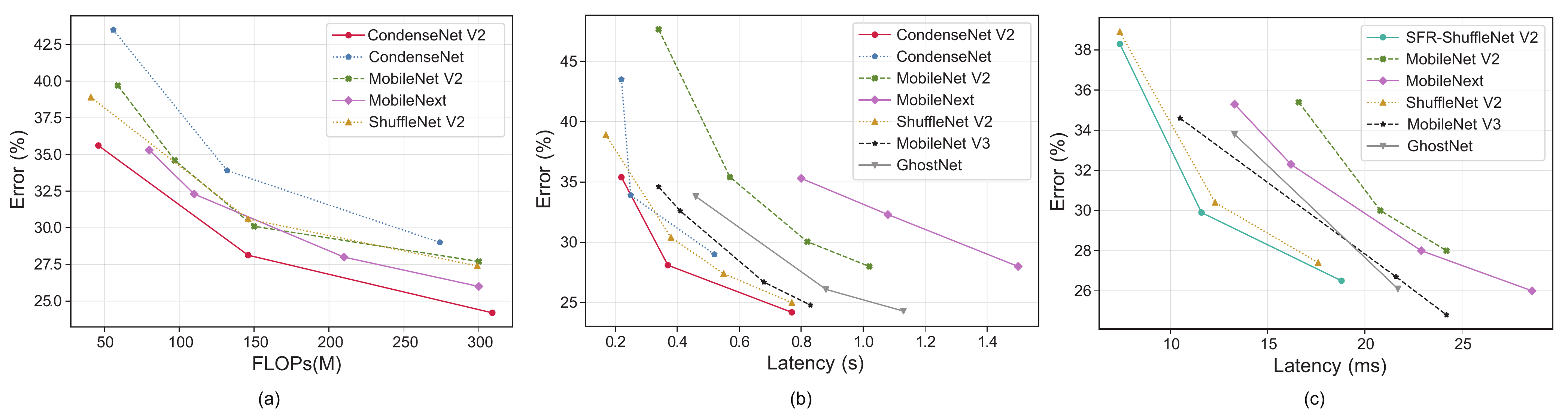}
\caption{(a) Top-1 err. \textit{v.s.} FLOPs on ImageNet. (b) Inference speed on the ARM processor. (c) Inference speed on iPhone XS Max.}
	\label{fig:flops}
\end{figure*}

We further compare our networks with several efficient network architectures designed by handcraft, including MobileNetV2~\cite{sandler2018inverted}, CondenseNet~\cite{huang2018condense}, MobileNeXt~\cite{zhou2020rethinking} and ShuffleNetV2\cite{ma2018shufflenet}. The results are shown in Figure \ref{fig:flops} (a). One can observe that the proposed \nameshort{} outperforms these models in terms of the computational efficiency.

In addition to the deep models in Figure \ref{fig:flops}, networks based on neural architecture search (NAS) are further compared with the proposed networks, including MobileNetV3 \cite{howard2019searching}, RegNetX~\cite{radosavovic2020designing}, ProxylessNAS\cite{cai2018proxylessnas} and MnasNet \cite{tan2018MnasNet}. The results are summarized in Table~\ref{tab:ImageNetResultAll}. We group different models according to their computational costs. Importantly, the proposed \nameshort{} does not leverage any technique of NAS, however, it can outperform most of the competitive baselines with similar FLOPs.

To show how our method can benefit from NAS, we further deploy our SFR module on ShuffleNetV2+\footnote{\url{https://github.com/megvii-model/ShuffleNet-Series/tree/master/ShuffleNetV2\%2B}}, a strengthened version of ShuffleNetV2 obtained by one-shot NAS based on ShuffleNet Units. The proposed SFR-ShuffleNetV2+ outperforms both MobileNet V3 and the original ShuffleNetV2+, which confirms the effectiveness of the proposed SFR module. More experiments on ImageNet are provided in Appendix B.

\begin{table}[t]\footnotesize
\caption{Comparison of Top-1 and Top-5 classification error rate (\%) with state-of-the-art efficient deep learning models on ImageNet. (L., M., and S. represent Large, Medium, Small, respectively.)}
\vspace{-5pt}
\begin{center}
\begin{tabular}{p{3.19cm}|p{0.7cm}<{\centering}|p{0.7cm}<{\centering}|  p{0.7cm}<{\centering}|p{0.7cm}<{\centering}}
    \toprule
\multirow{2}{*}{Model} & \multirow{2}{*}{FLOPs} & \multirow{2}{*}{Params} & Top-1& Top5  \\
& & & err.& err.\\
    \midrule
    \midrule
ShuffleNetV2 0.5$\times$~\cite{ma2018shufflenet} & 41M & 1.4M & 38.9 & 17.4 \\
0.4 MobileNetV2~\cite{sandler2018inverted} & 43M        & --  & 43.4 & -- \\
MobileNeXt-0.35~\cite{zhou2020rethinking} & 80M   & 1.8M  & 35.3 & -- \\

    \midrule
\nameshortm{}-A & 46M     & 2.0M  & \textbf{35.6} &  15.8\\
  \midrule
  \midrule
ShuffleNetV2 1.0$\times$~\cite{ma2018shufflenet} & 146M & 2.3M & 30.6 & 11.1 \\
0.75 MobileNetV2~\cite{sandler2018inverted} & 145M        & --  & 32.1 & -- \\
MobileNetV3 L. 0.75$\times$~\cite{howard2019searching}& 155M & 4.0M & 26.7& -- \\
RegNetX~\cite{radosavovic2020designing}& 200M&2.7M& 31.1&--\\
MobileNeXt-0.75~\cite{zhou2020rethinking} & 210M   & 2.5M  & 28.0 & -- \\
ShuffleNetV2+ S. & 156M  & 5.1M  & 25.9 & 8.3 \\
  \midrule
\nameshortm{}-B & 146M     & 3.6M  & 28.1 & 9.7\\
SFR-ShuffleNetV2+ S. & 161M  & 5.2M  & \textbf{25.5} & 8.2 \\
  \midrule
  \midrule
ShuffleNetV2 1.5$\times$~\cite{ma2018shufflenet} & 299M & -- & 27.4 & 9.4 \\
1.0 MobileNetV2~\cite{sandler2018inverted} & 300M & 3.4M & 28.0 & 9.0 \\
MobileNetV3 L. 1.0$\times$~\cite{howard2019searching}& 219M & 5.4M & 24.8& -- \\
RegNetX~\cite{radosavovic2020designing}& 400M&5.2M& 27.3&--\\
MobileNeXt-1.00~\cite{zhou2020rethinking} & 300M   & 3.4M  & 26.0 & -- \\
MnasNet-A1~\cite{tan2018MnasNet} & 312M   & 3.9M  & 24.8 & 7.5 \\
FE-Net 1.0$\times$~\cite{chen2019FENet}& 301M & 3.7M & 27.1 & -- \\
ESPNetV2~\cite{mehta2018espnetv2} & 284M & 3.5M & 27.9 & -- \\
ProxylessNAS\cite{cai2018proxylessnas}& 320M &4.1M&25.4&7.8\\
ShuffleNetV2+ M. & 222M  & 5.6M  & 24.3 & 7.4 \\
\midrule
\nameshortm{}-C & 309M     & 6.1M  & 24.1 & 7.3\\
SFR-ShuffleNetV2+ M. & 229M & 5.7M  & \textbf{23.9} &  7.3\\

    \bottomrule
\end{tabular}%
\end{center}
\label{tab:ImageNetResultAll}
\vspace{-25pt}	
\end{table}

\vspace{-10pt}	
\paragraph{Actual inference time.}
Since the proposed \nameshort{} is designed for edge devices, we further measure the actual inference speed of \nameshort{} on an ARM processor\footnote{Quad-Core ARM Cortex-A57 MPCore combined with Dual-Core NVIDIA Denver 2 64-Bit CPU.} and an iPhone XS Max (with Apple A12 Bionic). The single-thread mode with batch size 1 is used following \cite{howard2019searching} and we use a 224$\times$224 input image.

On the ARM processor, all models are implemented in PyTorch1.6.0. From Figure \ref{fig:flops} (b), one can observe that the proposed \nameshort{} achieves faster runtime under the same error compared with other light-weighted deep models. Specifically, our model obtains about 0.5\% lower top-1 error than MobileNetV3 with slightly lower latency. It is noteworthy that the power of the tested processor is lower than most smart mobile phones. We believe that such a speed test is necessary: although mobile phones and processors with high performance have been widely deployed and popularized nowadays, the computational resources of most edge devices, such as IoT products, are still highly limited. The proposed \nameshort{} outperforms most light-weighted networks in such a resource-limited scenario.

We further test the inference time on an iPhone XS Max, which can be considered as a high performance edge device. Our implementation is based on the Pytorch Mobile\footnote{\url{https://pytorch.org/mobile/home/.}}. The results are presented in Figure \ref{fig:flops} (c), from which we see that the SFR-ShuffleNetV2 outperforms other competitors. Here, only SFR-ShuffleNetV2 are tested because we found that the Pytorch Mobile might have poor support for group convolution operator: although \nameshort{}-A only has 46M FLOPs, its latency on iPhone is still up to 34.6ms.

\subsection{Experiments on CIFAR}
\paragraph{Implementation details.}
We apply SGD to train all the models with similar hyper-parameters setting as in ~\cite{huang2018condense}. We use the cosine learning rate annealing with an initial learning rate of 0.1. The training process lasts for 300 epochs with a mini-batch size of 64. Other training settings are the same as the experiments on ImageNet. \nameshort{}s on CIFAR follow the configuration listed below. The network consists of three dense blocks with the same number of layers, and the resolutions of feature maps are $32\!\times\!32$, $16\!\times\!16$, and $8\!\times\!8$, respectively. The growth rates are set to 8, 16, 32 for each block. The $C$, $S$, and $G$ are all set to 4. We modify the number of blocks $d$ in each stage to change the computational complexity of \nameshort{}. Moreover, we do not implement SE and HS in \nameshort{} for CIFAR models, and the last Conv2d 1$\times$1 is also removed.

\begin{table}[t]\footnotesize
\caption{Comparision of error rate (\%) with other state-of-the-art efficient models on CIFAR-10 and CIFAR-100.}
\label{tab:CIFARResult}
\begin{center}
\begin{tabular}{l|c c c c}
\toprule
Model & FLOPs & Params & C-10 & C-100\\
\midrule
\multicolumn{5}{c}{ResNet-based} \\
\midrule
CP~\cite{he2017channel} & 62M & - & 8.20 &-\\
PFEC~\cite{li2016pruning} & 90M & 0.73M & 6.94 &-\\
LECN~\cite{liu2017learning} & 124M & 1.21M & 5.27 & 23.91\\
NISP~\cite{yu2017nisp} & 142M & 0.96M & 6.88 & -\\
FPGM~\cite{he2018pruning} & 121M & - & 6.24 & -\\
\midrule
\multicolumn{5}{c}{DenseNet-based} \\
\midrule
LECN~\cite{liu2017learning} & 190M & 0.66M & 5.19 & 25.28\\
CondenseNet~\cite{huang2018condense} & 65M & 0.52M & 5.00 & 23.64\\
\midrule
\nameshortm{}-110 & 41M & 0.48M & 4.65 & 23.94 \\
\nameshortm{}-146 & 62M & 0.78M & \textbf{4.35} & \textbf{22.52} \\
\bottomrule
\end{tabular}%
\end{center}
\vspace{-15pt}	
\end{table}

\vspace{-5pt}
\paragraph{Results on CIFAR.}
We show the comparison results of \nameshortm{}s and other competitive baselines in Table \ref{tab:CIFARResult}. The baselines include several recently proposed network pruning algorithms. It can be observed that the \nameshortm{}s outperform all other approaches with lower error rates and less computational costs --- indicating that the effectiveness of the proposed feature reuse mechanisms.

\subsection{Experiments on MS COCO}

MS COCO~\cite{lin2014microsoft} is used for evaluating the generalization ability of our networks. Following \cite{ren2015fasterRCNN}, we use the \textit{trainval35k} split as training data and report the results in mean Average Precision (mAP) on \textit{minival split}. Faster R-CNN~\cite{ren2015fasterRCNN} with Feature Pyramid Networks (FPN)~\cite{lin2017featurePyramid} and RetinaNet~\cite{lin2017focal} are implemented as detection frameworks. Only backbone networks are replaced during experiments. Models are pretrained on ImageNet and then finetuned on the detection task. During finetuning, we train all models using SGD for 12 epochs. The input images are resized to a short side of 800 and a long side not exceed 1333.
The backbone FLOPs are calculated with 224$\times$224 input size following~\cite{ma2018shufflenet}.
The detection results are shown in Table~\ref{tab:coco}. As we can see, with comparable computational cost, our \nameshortm{}-C achieves higher mAP compared with ShuffleNetV2 and MobileNetV2, both on RetinaNet and Faster R-CNN frameworks.

\begin{table}[t]\small
\caption{Results on the MS COCO dataset.}
\vspace{6pt}
\label{tab:coco}
\begin{center}
\begin{tabular}{p{1.45cm}|p{3.1cm} |p{1.3cm}<{\centering} |p{0.6cm}<{\centering} }
\toprule
Detection Framework &Backbone  & Backbone FLOPs & mAP \\
\midrule
&ShuffleNetV2 0.5$\times$~\cite{ma2018shufflenet} & 41M & 22.1 \\
& \nameshort{}-A &46M &\textbf{23.5}\\
\cmidrule{2-4}
Faster&ShuffleNetV2 0.5$\times$~\cite{ma2018shufflenet} & 146M & 27.4 \\
R-CNN&\nameshort{}-B &146M &\textbf{27.9}\\
\cmidrule{2-4}
&MobileNetV2 1.0$\times$~\cite{sandler2018inverted} & 300M & 30.6\\
&ShuffleNetV2 1.5$\times$~\cite{ma2018shufflenet} & 299M & 30.2 \\
&SFR-ShuffleNetV2 1.5$\times$ & 306M & 30.7 \\
& \nameshort{}-C &309M &\textbf{31.4}\\
\midrule
\multirow{3}{*}{RetinaNet} &MobileNetV2 1.0$\times$~\cite{sandler2018inverted} & 300M & 29.7\\
&ShuffleNetV2 1.5$\times$ \cite{ma2018shufflenet} & 299M & 29.1 \\
& \nameshort{}-C & 309M &\textbf{31.7}\\
\bottomrule
\end{tabular}%
\end{center}
\vspace{-15pt}
\end{table}

\vspace{-5pt}
\section{Conclusion}
 \vspace{-5pt}
In this paper, we proposed a novel \textit{sparse feature reactivation} module, which can strategically reactivate a set of previous features to increase their utility for later layers. Importantly, the features to be reactivated are not \textit{pre-defined}, but \textit{learned} automatically during training. Due to the sparsity of the feature reactivation, this procedure can be highly computational-efficient. Therefore, the resulting model, \nameshort{}, based on the proposed SFR module can achieve high efficiency during inference. Encouraging results have been obtained on the image classification tasks (ImageNet and CIFAR) and the COCO object detection task, without resorting to neural architecture search.


\vspace{-5pt}
\section*{Acknowledgement}
 \vspace{-5pt}
This work is supported in part by the National Key R$\&$D Program of China (2020AAA0105200),  the National Natural Science Foundation of China (61906106, 62022048), the Institute for Guo Qiang of Tsinghua University and Beijing Academy of Artificial Intelligence. 

{\small
\bibliographystyle{ieee_fullname}
\bibliography{citations_sup}
}

\clearpage

\begin{appendix}
\section{Network Structure}
\subsection{Implementation details of SFR-ShuffleNetV2 and SFR-ShuffleNetV2+}

In this section, we provide more details for building SFR-ShuffleNetV2 and SFR-ShuffleNetV2+ based on ShuffleNetV2~\cite{ma2018shufflenet} and ShuffleNetV2+\footnote{\url{https://github.com/megvii-model/ShuffleNet-Series/tree/master/ShuffleNetV2\%2B}}, respectively. The general network architecture of SFR-ShuffleNetV2 and SFR-ShuffleNetV2+ are shown in Table \ref{tab:sfrshufflenetv2} and Table \ref{tab:sfrshufflenetv2plus}. The ShuffleNet Unit with stride 2 in Table \ref{tab:sfrshufflenetv2} is implemented following \cite{ma2018shufflenet}. The details of SFR-ShuffleNet Unit is illustrated in Figure 5 of the main paper. For SFR-ShuffleNetV2+, we follow the same principle to integrate the SFR modules into different ShuffleNet Units. Moreover, different from \nameshort{}, the SFR procedure is conducted by 3$\times$3 convolutions in SFR-ShuffleNet. There are four types of ShuffleNet Units in ShuffleNetV2+ (refer to the original implementation of ShuffleNetV2+ for more details). Note that we do not conduct the feature reactivation on the unit with a stride equal to 2. Additionally, the network configurations of SFR-ShuffleNetV2 and SFR-ShuffleNetV2+ are provided in Table \ref{tab:config-sfr-shufflenetv2} and \ref{tab:config-sfr-shufflenetv2plus}.

\begin{table}[h]\footnotesize
    \caption{Network architecture of SFR-ShuffleNetV2. The number of output channels and the sparse factor of $i$-th stage are $c_i$ and $S_i$, respectively. The number of units is denoted by $n$.}
    \label{tab:sfrshufflenetv2}
    \vspace{-5pt}
    \renewcommand{\arraystretch}{1.1}
    \begin{center}
    \begin{tabular}{c|c||c|c|c}
    \hline
    Output size & $c$ & Operator & $n$  & $S$ \\
    \hline
    $224\!\times\!224$ & 3 & Input Image & - & - \\
    \hline
    $112\!\times\!112$ & 24 &  Conv2d $3\!\times\!3$ (stride 2) & - & - \\
    $56\!\times\!56$ & 24 &  MaxPool $3\!\times\!3$ (stride 2) & - & - \\
    \hline
    \multirow{2}{*}{$28\!\times\!28$} & \multirow{2}{*}{$c_1$} & ShuffleNet Unit (stride 2) & 1 & - \\
    & & SFR-ShuffleNet Unit & 3 & $S_1$\\
    \hline
    \multirow{2}{*}{$14\!\times\!14$} & \multirow{2}{*}{$c_2$} & ShuffleNet Unit (stride 2) & 1 & - \\
    & & SFR-ShuffleNet Unit & 7 & $S_2$ \\
    \hline
    \multirow{2}{*}{$7\!\times\!7$} & \multirow{2}{*}{$c_3$} & ShuffleNet Unit (stride 2) & 1 & - \\
    & & SFR-ShuffleNet Unit & 3 & $S_3$ \\
    \hline
    $7\!\times\!7$ & 1024 & Conv2d $1\!\times\!1$ & - & -\\
    \hline
    $1\!\times\!1$ & - &  AvgPool $7\!\times\!7$ & - & -\\
    \hline
    $1\!\times\!1$ & 1000 & FC & - & -\\
    \hline
    \end{tabular}
    \end{center}
    \vspace{-5pt}
\end{table}

\begin{table}[h]\small
    \caption{Network configurations for SFR-ShuffleNetV2 models.}
    \vspace{5pt}
    \renewcommand{\arraystretch}{1.1}
    \centering
    \begin{tabular}{c|c|c}
    \toprule
        Network & $\{c_i\}$ & $\{S_i\}$ \\
    \midrule
        SFR-ShuffleNetV2 0.5x & 48-96-192 & 24-48-96 \\
        SFR-ShuffleNetV2 1.0x & 116-232-464 & 58-116-232 \\
        SFR-ShuffleNetV2 1.5x & 176-352-704 & 88-176-352 \\
    \bottomrule
    \end{tabular}
    \vspace{5pt}
    \label{tab:config-sfr-shufflenetv2}
\end{table}

\begin{table*}[t]\footnotesize
    \renewcommand{\arraystretch}{1.1}
    \begin{center}
    \caption{Network architecture of SFR-ShuffleNetV2+. The number of output channels and the sparse factor of $i$-th stage are $c_i$ and $S_i$, respectively. The number of units is denoted by$n$. SE and HS denote whether using Squeeze-Excitation~\cite{hu2017squeeze} and Hard-Swish module in this unit.}
    \label{tab:sfrshufflenetv2plus}
    \vspace{5pt}
    \begin{tabular}{p{1.3cm}<{\centering}|p{0.5cm}<{\centering}|p{4cm}<{\centering}|p{0.7cm}<{\centering}|p{0.7cm}<{\centering}|p{0.7cm}<{\centering}|p{0.7cm}<{\centering}}
    \hline
    Output size & $c$ & Operator & $n$ & $S$ & SE & HS \\
    \hline
    $112\!\times\!112$ & 16 & Conv2d $3\!\times\!3$ (stride 2) & 1 & - & - & 1 \\
    \hline
    \multirow{4}{*}{$56\!\times\!56$} & \multirow{4}{*}{$c_1$} & ShuffleNet $3\!\times\!3$ Unit (stride 2) & 1 & - & - & - \\
    & & SFR-ShuffleNet $3\!\times\!3$ Unit & 1 & $S_1$ & - & - \\
    & & SFR-ShuffleNet Xception Unit & 1 & $S_1$ & - & - \\
    & & SFR-ShuffleNet $5\!\times\!5$ Unit & 1 & $S_1$ & - & - \\
    \hline
    \multirow{3}{*}{$28\!\times\!28$} & \multirow{3}{*}{$c_2$} & ShuffleNet $5\!\times\!5$ Unit (stride 2) & 1 & - & - & 1 \\
    & & SFR-ShuffleNet $5\!\times\!5$ Unit & 1 & $S_2$ & - & 1 \\
    & & SFR-ShuffleNet $3\!\times\!3$ Unit & 2 & $S_2$ & - & 1 \\
    \hline
    \multirow{7}{*}{$14\!\times\!14$} & \multirow{7}{*}{$c_3$} & ShuffleNet $7\!\times\!7$ Unit (stride 2) & 1 & - & 1 & 1 \\
    & & SFR-ShuffleNet $3\!\times\!3$ Unit & 1 & $S_3$ & 1 & 1 \\
    & & SFR-ShuffleNet $7\!\times\!7$ Unit & 1 & $S_3$ & 1 & 1 \\
    & & SFR-ShuffleNet $5\!\times\!5$ Unit & 2 & $S_3$ & 1 & 1 \\
    & & SFR-ShuffleNet $3\!\times\!3$ Unit & 1 & $S_3$ & 1 & 1 \\
    & & SFR-ShuffleNet $7\!\times\!7$ Unit & 1 & $S_3$ & 1 & 1 \\
    & & SFR-ShuffleNet $3\!\times\!3$ Unit & 1 & $S_3$ & 1 & 1 \\
    \hline
    \multirow{4}{*}{$7\!\times\!7$} & \multirow{4}{*}{$c_4$} & ShuffleNet $7\!\times\!7$ Unit (stride 2) & 1 & $S_4$ & 1 & 1 \\
    & & SFR-ShuffleNet $5\!\times\!5$ Unit & 1 & $S_4$ & 1 & 1 \\
    & & SFR-ShuffleNet Xception Unit & 1 & $S_4$ & 1 & 1 \\
    & & SFR-ShuffleNet $7\!\times\!7$ Unit & 1 & $S_4$ & 1 & 1 \\
    \hline
    $7\!\times\!7$ & 1280 & Conv2d $1\!\times\!1$ & - & - & - & 1 \\
    \hline
    $1\!\times\!1$ & - &  AvgPool $7\!\times\!7$ & - & - & - & - \\
    \hline
    $1\!\times\!1$ & 1280 & SE Module & - & - & 1 & - \\
    \hline
    $1\!\times\!1$ & 1280 & FC & - & - & - & 1 \\
    \hline
    $1\!\times\!1$ & 1000 & FC & - & - & - & -\\
    \hline
    \end{tabular}
    \end{center}
\end{table*}

\begin{table}[h]\small
    \caption{Network configurations for SFR-ShuffleNetV2+ models. S. and M. represent Small and Medium, respectively.}
    \vspace{5pt}
    \renewcommand{\arraystretch}{1.1}
    \centering
    \begin{tabular}{c|c|c}
    \toprule
        Network & $\{c_i\}$ & $\{S_i\}$ \\
    \midrule
        SFR-ShuffleNetV2+ S. & 36-104-208-416 & 18-52-104-213 \\
        SFR-ShuffleNetV2+ M. & 48-128-256-512 & 24-64-128-256 \\
    \bottomrule
    \end{tabular}
    \vspace{5pt}
    \label{tab:config-sfr-shufflenetv2plus}
\end{table}

\subsection{Implementation details of CondenseNet}
In CondenseNet~\cite{huang2018condense}, only the network architecture of CondenseNet-C with 300M FLOPs is provided. In order to conduct a more comprehensive comparison, we further design CondenseNet-A/B which are under another two computation levels(50M and 150M). The general network architecture and configurations of CondenseNet are provided in Table \ref{tab:condensenet} and Table \ref{tab:config-condensenet}, respectively. Here, the DenseLayers in Table \ref{tab:condensenet} are implemented with learned group convolutions.

\begin{table}[t]\footnotesize
    \caption{Network architecture of CondenseNet. The number of layers and the growth rate of $i$-th dense block are $d_i$ and $k_i$, respectively.}
    \vspace{-5pt}
    \renewcommand{\arraystretch}{1.1}
    \begin{center}
    \begin{tabular}{c||c|c|c}
    \hline
    Input & Operator & $d$ & $k$ \\
    \hline
    $224\!\times\!224$ &  Conv2d $3\!\times\!3$ (stride 2)& - & -\\
    \hline
    $112\!\times\!112$ & DenseLayer & $d_1$ & $k_1$   \\
    $112\!\times\!112$ & AvgPool $2\!\times\!2$ (stide 2)&- & -\\
    \hline
    $56\!\times\!56$ & DenseLayer &$d_2$ & $k_2$\\
    $56\!\times\!56$ & AvgPool $2\!\times\!2$ (stride 2)& - & - \\
    \hline
    $28\!\times\!28$ & DenseLayer &$d_3$ & $k_3$ \\
    $28\!\times\!28$ & AvgPool $2\!\times\!2$ (stride 2)  & - & - \\
    \hline
    $14\!\times\!14$ & DenseLayer & $d_4$ & $k_4$ \\
    $14\!\times\!14$ & AvgPool $2\!\times\!2$ (stride 2) & - & -\\
    \hline
    $7\!\times\!7$ & DenseLayer &$d_5$ & $k_5$\\
    \hline
    $1\!\times\!1$ & AvgPool $7\!\times\!7$ & - & - \\
    \hline
    $1\!\times\!1$ & FC & - & -\\
    \hline
    \end{tabular}
    \end{center}
    \vspace{-5pt}
    \label{tab:condensenet}
\end{table}

\begin{table}[t]\small
    \caption{Detailed network configurations for CondenseNet models. $C$ denotes the condense factor for learned group convolution.}
    \vspace{5pt}
    \renewcommand{\arraystretch}{1.1}
    \centering
    \begin{tabular}{c|c|c|c}
    \toprule
        Network & $\{d_i\}$ & $\{k_i\}$ & $C$ \\
    \midrule
        CondenseNet-A & 2-4-6-8-4 & 8-8-16-32-64 & 8\\
        CondenseNet-B & 2-4-6-8-6 & 6-12-24-48-96 & 6 \\
        CondenseNet-C & 4-6-8-10-8 & 8-16-32-64-128 & 8 \\
    \bottomrule
    \end{tabular}
    \label{tab:config-condensenet}
\end{table}

\section{A comperhensive study of efficient deep learning models on ImageNet}
In this section, we provide more comprehensive comparisons between the proposed network architectures and other state-of-the-art efficient deep learning models. These models are grouped into three levels of computational costs, including 50M, 150M and 300M FLOPs. Our comparisons include the most efficient network architectures: (1) Handcrafted Light-weighted CNN architectures, such as CondenseNet~\cite{huang2018condense}, MobileNetV1~\cite{howard2017mobilenets}, MobileNetV2~\cite{sandler2018inverted}, ShuffleNetV1~\cite{zhang2017shufflenet}, ShuffleNetV2~\cite{ma2018shufflenet}, IGCV3~\cite{sun2018igcv3}, Xception~\cite{chollet2016xception} and ESPNetV2~\cite{mehta2018espnetv2}, are shown in Table \ref{tab:ImageNetResultAll-Hand}. 2) NAS based methods, such as NASNet~\cite{zoph2017learning}, PNASNet~\cite{liu2018progressive}, MnasNet~\cite{tan2018MnasNet}, ProxylessNas~\cite{cai2018proxylessnas}, AmoebaNet~\cite{real2019regularized}, GhostNet~\cite{han2020GhostNet}, MobileNetV3~\cite{howard2019searching} and RegXNet~\cite{radosavovic2020designing}, are shown in Table \ref{tab:ImageNetResultAll-NAS}.

From the results in Table~\ref{tab:ImageNetResultAll-Hand}, we conclude that the proposed \nameshortm{} are superior to many other handcraft-designed efficient deep CNNs significantly. We also observe that the proposed networks outperform the CondenseNets by a large margin, which demonstrates that the effectiveness of our SFR module.

As we can see, in Table \ref{tab:ImageNetResultAll-NAS}, our \nameshortm{}-C surpass most of the efficient models based on NAS under the computational budget of $\sim$300M. Note that our \nameshortm{}-C's FLOPs is only half of NASNet-A, AmoebaNet-C and PNASNet-5, however, \nameshortm{}-C still outperform these NAS based models. Although the EfficientNetB0 achieves the best performance when the computational budget is $\sim$300M, its FLOPs is much larger than \nameshortm{}-C (390M vs 309M). The MobileNetV3 outperforms our models with $\sim$50M and $\sim$150M FLOPs which can be due to the effectiveness of NAS. Since our implemented SFR-ShuffleNetV2+ Small can surpass the MobileNetV3 Large 0.75 $\times$ by 1.2 percent in terms of Top-1 Error, we believe that \nameshortm{}'s performance can be further boosted by NAS algorithms. Therefore, the future work will mainly focus on applying NAS methods on the proposed \nameshortm{}.



\begin{table*}[ht]\small
    \caption{Comparison of Top-1 and Top-5 classification error rate (\%) with state-of-the-art handcraft-designed efficient models on the ILSVRC validation set.}
    \begin{center}
    \begin{tabular}{p{4.3cm}|p{1.5cm}<{\centering}p{1.5cm}<{\centering}  p{1.5cm}<{\centering} p{1.5cm}<{\centering}}
        \toprule
    Model & FLOPs & Params & Top-1 err. & Top-5 err. \\
        \midrule
        \midrule
    ShuffleNetV1 0.5$\times$(g=3)~\cite{zhang2017shufflenet}& 38M & 1.0M & 41.2 & 19.0 \\
    MobileNetV1-0.25~\cite{howard2017mobilenets}& 41M & -- & 49.4 & -- \\
    ShuffleNetV2 0.5$\times$~\cite{ma2018shufflenet} & 41M & 1.4M & 38.9 & 17.4 \\
    MobileNetV2-0.40~\cite{sandler2018inverted} & 43M & -- & 43.4 & -- \\
    CondenseNet-A~\cite{huang2018condense}& 56M & 0.9M & 43.5 & 20.2 \\
    \midrule
    SFR-ShuffleNetV2 0.5$\times$ & 43M & 1.4M & 38.3 & 17.0 \\
    \nameshortm{}-A & 46M & 2.0M & \textbf{35.6} & \textbf{15.8}\\
    \midrule
    \midrule
    MobileNeXt-0.50~\cite{zhou2020rethinking} & 110M & 2.1M & 32.3 & -- \\
    ShuffleNetV1 1.0$\times$(g=3)~\cite{zhang2017shufflenet}& 138M & 1.9M & 32.2 & 12.3 \\
    MobileNetV1-0.50~\cite{howard2017mobilenets}& 149M & -- & 36.3 & -- \\
    ShuffleNetV2 1.0$\times$~\cite{ma2018shufflenet} & 146M & 2.3M & 30.6 & 11.1 \\
    MobileNetV2-0.75~\cite{sandler2018inverted} & 145M & --  & 32.1 & -- \\
    IGCV3-D-0.70~\cite{sun2018igcv3}& 210M & 2.8M & 31.5 & --\\
    CondenseNet-B~\cite{huang2018condense} & 132M & 2.1M & 33.9 & 13.1 \\
    \midrule
    SFR-ShuffleNetV2 1.0$\times$ & 150M & 2.3M & 29.9 & 10.9 \\
    \nameshortm{}-B & 146M & 3.6M & \textbf{28.1} & \textbf{9.7}\\
    \midrule
    \midrule
    Xception 1.5$\times$~\cite{chollet2016xception} & 305M & -- & 29.4 & -- \\
    ShuffleNetV1 1.5$\times$~\cite{zhang2017shufflenet} & 292M & -- & 28.5 & -- \\
    MobileNetV1-0.75~\cite{howard2017mobilenets}& 325M & -- & 31.6 & -- \\
    ShuffleNetV2 1.5$\times$~\cite{ma2018shufflenet} & 299M & -- & 27.4 & -- \\
    MobileNetV2-1.00~\cite{sandler2018inverted} & 300M & 3.4M & 28.0 & -- \\
    MobileNeXt-1.00~\cite{zhou2020rethinking} & 300M & 3.4M & 26.0 & -- \\
    IGCV3-D-1.00~\cite{sun2018igcv3}& 318M & 3.5M & 27.8 & --\\
    FE-Net 1.0$\times$~\cite{chen2019FENet}& 301M & 3.7M & 27.1 & -- \\
    ESPNetV2~\cite{mehta2018espnetv2} & 284M & 3.5M & 27.9 & -- \\
    CondenseNet-C~\cite{huang2018condense} & 274M & 2.9M & 29.0 & 10.0\\
    \midrule
    SFR-ShuffleNetV2 1.5$\times$ & 306M & 3.5M & 26.5 & 8.6 \\
    \nameshortm{}-C & 309M & 6.1M & \textbf{24.1} & \textbf{7.3}\\
    \bottomrule
    \end{tabular}%
    \end{center}
    \label{tab:ImageNetResultAll-Hand}
\end{table*}

\begin{table*}[ht]\small
    \caption{Comparison of Top-1 and Top-5 classification error rate (\%) with state-of-the-art NAS based efficient models on the ILSVRC validation set.}
    \begin{center}
    \begin{tabular}{p{4.3cm}|p{1.5cm}<{\centering}p{1.5cm}<{\centering}  p{1.5cm}<{\centering} p{1.5cm}<{\centering}}
    \toprule
    Model & FLOPs & Params & Top-1 err. & Top-5 err. \\
    \midrule
    \midrule
    MobileNetV3 Large 0.75$\times$~\cite{howard2019searching}& 155M & 4.0M & 26.7& -- \\
    RegNetX~\cite{radosavovic2020designing}& 200M&2.7M& 31.1&--\\
    GhostNet 1.0$\times$~\cite{han2020GhostNet}& 141M&5.2M& 26.1& 8.6 \\
    \midrule
    SFR-ShuffleNetV2+ Small & 161M & 5.2M & \textbf{25.5} & \textbf{8.2}\\
    \midrule
    \midrule
    MobileNetV3 Large 1.00$\times$~\cite{howard2019searching}& 219M & 5.4M & 24.8& -- \\
    GhostNet 1.3$\times$~\cite{han2020GhostNet}& 226M&7.3M& 24.3& 7.3 \\
    MnasNet-A1~\cite{tan2018MnasNet} & 312M & 3.9M & 24.8 & 7.5 \\
    ProxylessNAS\cite{cai2018proxylessnas}& 320M & 4.1M & 25.4 & 7.8 \\
    RegNetX~\cite{radosavovic2020designing}& 400M & 5.2M & 27.3 &-- \\
    NASNet-A~\cite{zoph2017learning}& 564M & 5.2M & 26.0 & 8.4 \\
    AmoebaNet-C~\cite{real2019regularized}& 570M & 6.4M & 24.3 & 7.6 \\
    PNASNet-5~\cite{liu2018progressive}& 588M & 5.1M & 25.8 & 8.1 \\
    \midrule
    SFR-ShuffleNetV2+ Medium & 229M & 5.7M & \textbf{23.9} & \textbf{7.3}\\
    \bottomrule
    \end{tabular}%
    \end{center}
    \label{tab:ImageNetResultAll-NAS}
\end{table*}

\section{Experimental Setup on ImageNet}
In our experiments, all our models are conducted under Pytorch Implementation~\cite{paszke2019pytorch}. Our training is based on the open-source code\footnote{\url{https://github.com/rwightman/pytorch-image-models/}} which successfully reproduces the reported performance in MobileNetV3~\cite{howard2019searching}. We follow most of the training settings used in MobileNetV3~\cite{howard2019searching}, except that we use the stochastic gradient descent (SGD) optimizer with an initial learning rate of 0.4, the cosine learning rate~\cite{loshchilov2016sgdr}, a Nesterov momentum of weight 0.9 without dampening and a weight decay of $4\!\times\!10^{-5}$ when batch size is 1024.



\end{appendix}

\end{document}